\documentclass[10pt,journal,compsoc]{IEEEtran}
\usepackage{graphicx}
\usepackage{amsmath}
\usepackage{algorithm, algorithmic}
\usepackage{multirow}
\usepackage{booktabs}
\usepackage{algorithm, algorithmic}
\usepackage{threeparttable}
\usepackage{graphics}
\usepackage{sidecap}

\usepackage{subcaption}
\usepackage{color}
\usepackage{hologo} 
\usepackage{url}
\usepackage{booktabs} 
\usepackage{tablefootnote} 

\ifCLASSOPTIONcompsoc
  \usepackage[nocompress]{cite}
\else
  \usepackage{cite}
\fi
\ifCLASSINFOpdf
\else
\fi
\begin{document}
%
\title{Bare Demo of IEEEtran.cls for\\ IEEE Computer Society Journals}
%
%
%
%

\author{Michael~Shell,~\IEEEmembership{Member,~IEEE,}
        John~Doe,~\IEEEmembership{Fellow,~OSA,}
        and~Jane~Doe,~\IEEEmembership{Life~Fellow,~IEEE}
\IEEEcompsocitemizethanks{\IEEEcompsocthanksitem M. Shell was with the Department
of Electrical and Computer Engineering, Georgia Institute of Technology, Atlanta,
GA, 30332.\protect\\
E-mail: see http://www.michaelshell.org/contact.html
\IEEEcompsocthanksitem J. Doe and J. Doe are with Anonymous University.}
\thanks{Manuscript received April 19, 2005; revised August 26, 2015.}}

%
%

\markboth{IEEE Transactions on Neural Networks and learning systems}%
{Shell \MakeLowercase{\textit{et al.}}: Bare Demo of IEEEtran.cls for Computer Society Journals}
%


\title{OpenAPMax: Abnormal Patterns-based Model for Real-World Alzheimer's Disease Diagnosis}

\author{Yunyou Huang, Xianglong Guan, Xiangjiang Lu, Xiaoshuang Liang, Xiuxia Miao, Jiyue Xie, Wenjing Liu, Li Ma, Suqin Tang, Zhifei Zhang, and Jianfeng Zhan
\IEEEcompsocitemizethanks{\IEEEcompsocthanksitem Y. Huang, X. Lu, X. Liang, X. Miao, J. Xie, W. Liu, and S. Tang are with the Key Lab of Education Blockchain and Intelligent Technology, Ministry of Education, and the Guangxi Key Laboratory of Multi-Source Information Mining and Security, Guangxi Normal University.\protect
\IEEEcompsocthanksitem X. Guan is with the School of Electronic and Information Engineering \& School of Integrated Circuits, Guangxi Normal University, Guilin 530015, China. He is also with the Key Lab of Education Blockchain and Intelligent Technology, Ministry of Education, Guangxi Normal University.

\IEEEcompsocthanksitem J. Zhan is with the State Key Laboratory of Computer Architecture, Institute of Computing Technology, Chinese Academy of Sciences, Beijing, 100086, China.

\IEEEcompsocthanksitem Z. Zhang is with the Department of Physiology and Pathophysiology, Capital Medical University, Beijing, 100069, China.

\IEEEcompsocthanksitem L. Ma is with the Guilin Medical University, Guilin 541001, China. She is also with the Guangxi Key Lab of Multi-Source Information Mining \& Security.}

\thanks{Correspondence authors: Jianfeng Zhan(zhanjianfeng@ict.ac.cn) or Zhifei Zhang(zhifeiz@ccmu.edu.cn) or Suqin Tang(sqtang@gxnu.edu.cn)}}

\IEEEtitleabstractindextext{%
\begin{abstract}
Alzheimer's disease (AD) cannot be reversed, but early diagnosis will significantly benefit patients' medical treatment and care. In recent works, AD diagnosis has the primary assumption that all categories are known a prior---a closed-set classification problem, which contrasts with the open-set recognition problem. This assumption hinders the application of the model in natural clinical settings. Although many open-set recognition technologies have been proposed in other fields, they are challenging to use for AD diagnosis directly since 1) AD is a degenerative disease of the nervous system with similar symptoms at each stage, and it is difficult to distinguish from its pre-state, and 2) diversified strategies for AD diagnosis are challenging to model uniformly. In this work, inspired by the concerns of clinicians during diagnosis, we propose an open-set recognition model, OpenAPMax, based on the anomaly pattern to address AD diagnosis in real-world settings. OpenAPMax first obtains the abnormal pattern of each patient relative to each known category through statistics or a literature search, clusters the patients' abnormal pattern, and finally, uses extreme value theory (EVT) to model the distance between each patient's abnormal pattern and the center of their category and modify the classification probability. We evaluate the performance of the proposed method with recent open-set recognition, where we obtain state-of-the-art results.
\end{abstract}

}

\maketitle

\IEEEdisplaynontitleabstractindextext

%
\IEEEpeerreviewmaketitle

\IEEEraisesectionheading{\section{Introduction}\label{sec:introduction}}

%
%
%
%
\IEEEPARstart{A}{lzheimer's} Disease (AD) is an incurable disease that causes significant harm to patients. Currently, the number of AD patients is approximately 50 million, increasing dramatically with societal ageing~\cite{livingston2020dementia}. Although AD diagnosis does not affect the disease's cure, the correct diagnosis will lead to proper nursing and treatment to alleviate the patient's symptoms, thereby significantly reducing the burden of the disease~\cite{anderson2019state}. Artificial intelligence technology is considered one of the most promising technologies for improving medical services and has been introduced into many medical fields, including Alzheimer's disease~\cite{tanveer2020machine,mahajan2020machine}.

Currently, many AI models are proposed for AD diagnosis to enhance the quality of medical services and expand AD screening. Zeng et al. adopted the traditional machine learning method and proposed a new switching delayed particle swarm optimization (SDPSO) algorithm to optimize the 
SVM  parameters, and then, the optimized SVM was used to diagnose Alzheimer's disease\cite{zeng2018new}. Khan et al. adopted a deep learning method and used transfer learning to build a VGG architecture for classification\cite{khan2019transfer}. To improve the generalization of the model, Lu et al. used transfer learning to construct a deep convolutional neural network ResNet model for classification on a large-scale dataset\cite{lu2022practical}. 
These methods 
can achieve good results in the classification and diagnosis of Alzheimer's disease in a closed-set environment. However, all of the recent works consider AD diagnosis as a classification problem in the closed world rather than an open-set recognition problem in the real world, which makes it impossible to implement the AI model in clinical settings.


\begin{figure}
\centering
\includegraphics[width=1.0\linewidth]{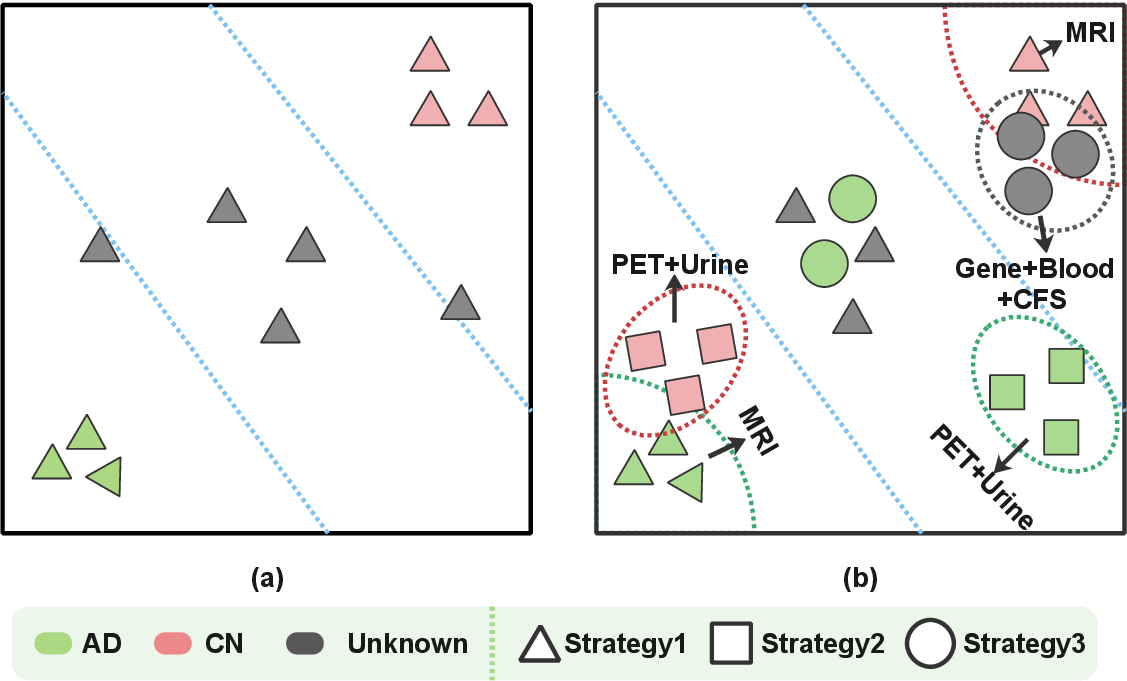}
\caption{Comparison of the open-set recognition task. (a) In the regular open-set recognition task, the data type and length of all samples are the same, and data in the same category have a similar distribution. (b) In open-set recognition of the diagnosis task, the data type and data length of different samples may be different, and the data distribution of samples in the same category may be very different. However, the data distribution of samples of different categories may be very similar.\label{task}}
\end{figure}

To make the AI model applicable in the real world, a large number of open-set recognition technologies have been proposed in various fields. Hitham et al. used the peak side ratio to characterize the posterior probabilities generated by a Platt-calibrated support vector machine (SVM). Then they used thresholds determined by this process to classify open audio data\cite{jleed2020open}. Mehadi et al. proposed a dual-loss neural network model that classifies open datasets by determining the distance between different instances\cite{hassen2020learning}, and Abhijit et al. proposed a new OpenMax layer that applies meta-recognition to the activation layer, improving the performance of the network in an open-set environment\cite{bendale2016towards}.

Open-set recognition technologies are roughly divided into discriminative models and generative models~\cite{geng2020recent}. 
Discriminative models generally train a classifier model in a closed set first and then observe the distribution of the classifier's output or intermediate results on the training set, observe the distribution of the intermediate results, or transform the output or intermediate results to another form to observe their distribution.
 Finally, it modifies the output of the classifier according to the distribution. However, as shown in Figure~\ref{task}(b), AD is similar to many other diseases (such as MCI). In a clinical setting, an AD diagnosis takes 2.7 years on average~\cite{kua2014natural}, which means that during the diagnosis process, AD may be indistinguishable from many very similar diseases, and it is difficult to distinguish unknown categories by observing the distribution of the models' output or the intermediate results.


The generative model mainly recognizes samples of unknown classes through reconstruction loss or data augmentation.~\cite{geng2020recent,kong2021opengan}. The model's key is whether the generator can fit the distribution of known samples or samples of unknown categories.~\cite{creswell2018generative}. However, as shown in Figure~\ref{task}(b), this goal is different from current research because each sample is composed of the same one or more types of data, and the type and amount of data for each sample may be different based on different diagnostic strategies. 

For example, diagnosing an AD patient $p_A$ in the training set may involve a combination strategy that includes a cognitive examination and MRI scanning, while diagnosing another AD patient $p_B$ in the training set may involve a combination strategy that includes a cognitive examination and a genetic examination. Furthermore, AD patient $p_C$ in the test set may have been diagnosed using a combination strategy that does not appear in the training set.
 
It is challenging to fit the sample's distribution based on the generation model due to the diversity of diagnostic methods; not only are the data of different categories of subjects distributed differently, but the data of patients in the same category are also distributed differently.

Both the discriminative model and generation model seek to classify patients with similar data into the same category. However, while investigating AD diagnosis in real clinical settings, we found that at the beginning of diagnosis, clinicians usually pay more attention to whether a patient's indicator is abnormal rather than the similarity between the patient's data and the data of the standard patient. In this work, inspired by the above phenomenon, we propose the abnormal pattern-based open-set recognition, named OpenAPMax, to diagnose AD patients in real-world settings. First, we selected indicators related to Alzheimer's disease through literature research. All of these indicators are only included in the basic information obtained through consultation and the information from routine cognitive tests by the outpatient department. Second, the patient's indicator value is compared with the normal value of each known category to obtain the patient's abnormal pattern. Third, a classifier that is able to distinguish AD from CN is trained. Finally, the extreme value theory (EVT) is utilized to model the distance between each patient's abnormal pattern and the center of their category and modify the classification probability.

To summarize, our contributions are as follows:

 \begin{itemize}
\item[(1)] Different from directly using the clinical data of subjects for open-set recognition, we introduce subjects' abnormal patterns into the model by imitating the clinician's concerns. This approach might inspire a rethinking of clinician behavior in real-world settings. To the best of our knowledge, this work is the first to use abnormal patterns in diagnosis based on open-set recognition.

\item[(2)] OpenAPMax can be integrated into other models without changing the model's architecture. Compared to OpenMax\cite{bendale2016towards} and OVRN\cite{jang2022collective}, OpenAPMax can provide better performance in complex diagnosis tasks.

\item[(3)] The experiments show that the model with OpenAPMax performs best in 
clinical settings with different complexities.

\end{itemize}

\section{Related Work}
\textbf{AD diagnosis model.} Early and accurate diagnosis of Alzheimer's disease allows patients to receive timely treatment and slows the progression of symptoms\cite{seltzer2004efficacy}. Li et al.~\cite{li2018alzheimer} proposed a classification method based on multi-cluster dense convolutional neural networks for AD diagnosis. This method can jointly learn features and classification without domain expert knowledge. Scholar et al.~\cite{vaithinathan2019novel} proposed a new AD classification method that extracts features from defined regions of interest and performs texture analysis, and Basaia et al.~\cite{basaia2019automated} used a single cross-sectional brain structure MRI scan to predict Alzheimer's disease based on a convolutional neural network classification approach. Liu et al.~\cite{liu2018joint} proposed a deep multitask multichannel learning framework that identifies discriminative anatomical landmarks from MR images and then extracts multiple image patches around these detected landmarks.

\textbf{Open-set Recognition.} To address the real-world problem, many open-set recognition technologies have been proposed and are divided into discriminative models and generative models~\cite{geng2020recent}. 
Alina et al.~\cite{9304605} combined
 the current closed-set models with multiple novelty detection strategies employed in general action classification to develop a model that recognizes 
previously unseen classifier behaviors  and proposed a new OpenDrive \& Act benchmark. Wentao et al.~\cite{Bao_2021_ICCV} proposed a new deep-evidence action recognition method that recognizes actions in an open testing set.
They used an 
evidence-based neural network
 to predict class-wise evidence from a given video, forming a Dirichlet distribution that determines the multiclass probability and prediction uncertainty of the input. Rafael et al.~\cite{vareto2017towards} combined hash functions and classification methods to estimate when probe samples are known. Specifically, they combined a Partial Least Squares Network and a fully connected network to generate an open-set face recognition classification model.
Ryota et al.~\cite{Yoshihashi_2019_CVPR} developed a classifier for classification reconstruction learning for open-set recognition using latent representation learning. This classifier enables robust unknown detection without compromising known classification accuracy.

\section{AD Diagnosis}
The subject dataset is denoted by $D$. Let $(X_i, y_i)$ denote the $ith$ sample in dataset $D$, where $X_i=\{x_1, x_2, ..., x_k\}$ is the subject's clinical data (obtained by enquiring about the subject or conducting a clinical examination), and $y_i$ is the subject's label.

Open-set recognition can not only classify the categories that have appeared according to the input but can also correctly judge the categories that do not appear in the input (unknown categories).
There are many forms of AD diagnosis based on open-set recognition, one of which can be transformed to minimize the objective function $L_{o}(W)$ as shown in Equation~\ref{open_set}.

\begin{equation}
\left\{
\begin{aligned}
  \label{open_set}
  &L_{o}(W)=\sum_{i=1}^n{l1(\hat{y_i}, y_{i})} +l2(W) \\
  &\hat{y_i}=f(W(X_i))\\
  &\hat{y_{i\_un}}=1-\sum_{j=1}^2\hat{y_i}[j]\\
\end{aligned}
\right.    
\end{equation}

$f$ is a score modifier based on EVT~\cite{bendale2016towards}, and $\hat{y_{i\_un}}$ is the probability that the sample belongs to an unknown category.

\section{Methods}

\subsection{The proposed model}
\begin{figure*}
\centering
\includegraphics[width=0.8\linewidth]{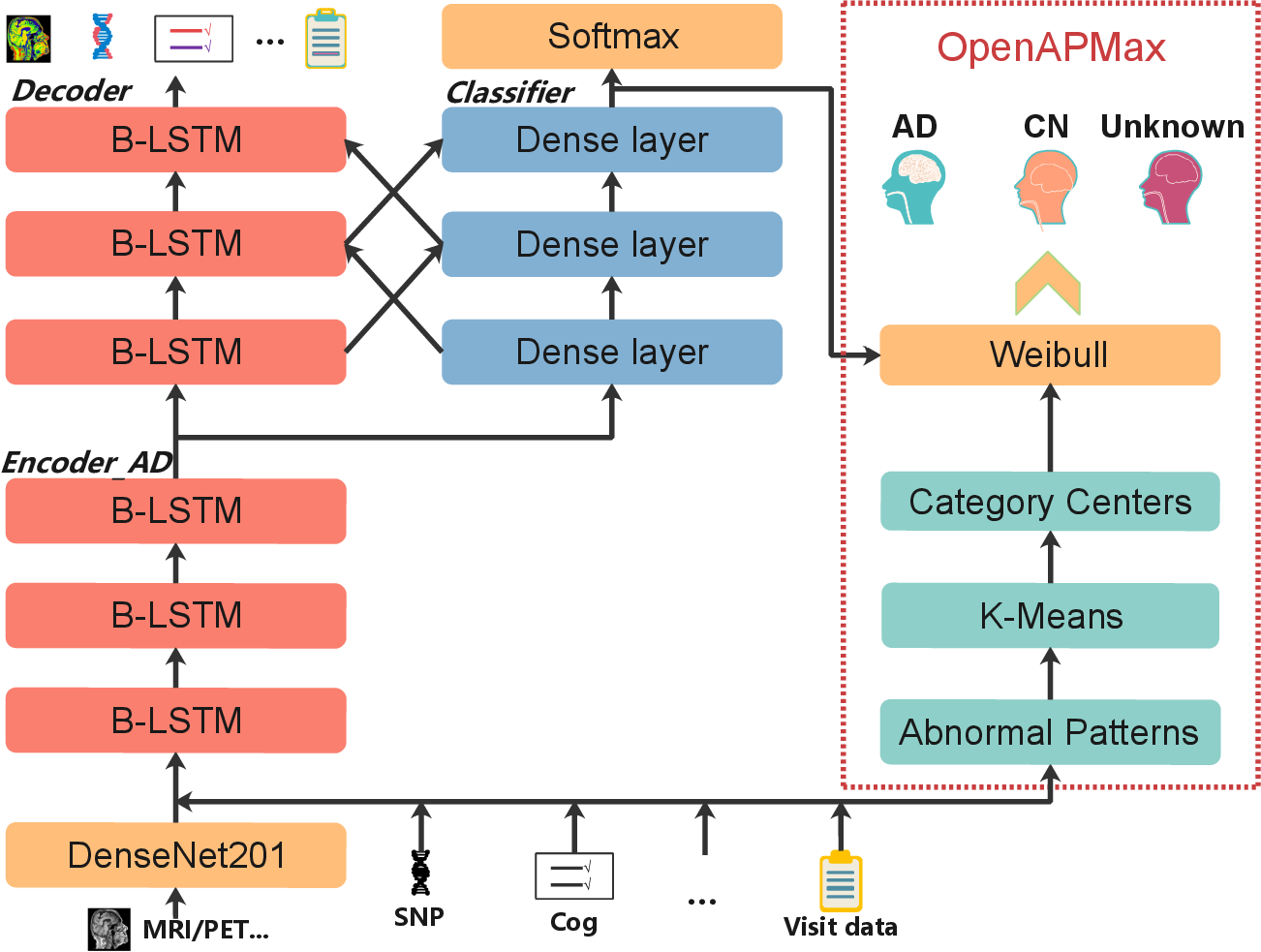}
\caption{Overview of the proposed model with OpenAPMax. The architecture accepts clinical data of unfixed type and length as input. The principal building blocks are an encoder, a decoder, a classifier, and an OpenAPMax layer. The OpenAPMax layer replaces the softmax layer during the prediction. \label{new_model}}
\end{figure*}

The framework of the AD diagnosis model with OpenAPMax consists of four components: AD diagnosis clinical data encoder $Encoder$, clinical data decoder $Decoder$, $Classifier$, and probability corrector $OpenAPMax$, as shown in Fig. ~\ref{new_model}.

The $Encoder$ consists of 3 layers of bidirectional LSTM (B-LSTM), and it is able to accept variable-length input and merge all of the clinical data since the length of that data will change with the change in diagnosis strategy. Note that all types of clinical image data will be extracted into vectors using the pretrained model, DenseNet201, before being input into the model~\cite{huang2017densely}. The $Decoder$ and $Encoder$ structures are similar and used to reconstruct data; the $Decoder$ is able to help the $Classifier$ retain more features that are irrelevant to AD and CN classification and improve the recognition performance for unknown category samples in the open environment. The $Classifier$ consists of three dense layers and a softmax layer, which is used to classify subjects in a closed setting. To enable the model to diagnose Alzheimer's disease in the real world, an open-set recognition mechanism based on abnormal patterns $OpenAPMax$ replaces the softmax layer after model training.

\subsection{OpenAPMax}

The deep learning network can be regarded as a feature extractor, and the output of the AV layer can be regarded as a characteristic of the sample. However, as the cause of Alzheimer's disease has not yet been determined, and because it is highly similar to many neurological diseases, it is difficult to distinguish between Alzheimer's disease and related diseases due to their overlapping characteristics. As shown in Figure~\ref{new_model}, we introduce the abnormal pattern to modify the open-set recognition mechanism. First, we selected 14 indicators related to the diagnosis of Alzheimer's disease according to the diagnostic guidelines for Alzheimer's disease~\cite{jack2011introduction,sperling2011toward,albert2011diagnosis,donohue2014preclinical}. We have noticed that all indicators are only obtained through inquiry or conventional cognitive testing because different patients usually receive different diagnostic strategies; we can only select the part where all patient data overlap. Second, for each category, the normal value range of each indicator is calculated as the basis for determining whether each indicator is normally relative to this category for each patient. The abnormal pattern is a sequence of 0 and 1
: 0 represents a normal indicator, and 1 represents an abnormal indicator. Third, as Algorithm~\ref{alg2} shows, OpenAPMax divides the abnormal patterns of each category into subcategories using a clustering algorithm since a disease usually has multiple subtypes; then, it obtains centers for the category by averaging. The Weibull distribution is used to model the distance between the abnormal pattern and the center of the category.

\begin{algorithm}
         \renewcommand{\algorithmicrequire}{\textbf{Input:}}
         \renewcommand{\algorithmicensure}{\textbf{Output:}}
         \caption{\textbf{OpenAPMax algorithm.}\label{alg2}}
         \begin{algorithmic}[1]
                   \REQUIRE The abnormal pattern dataset $X$, the FitHigh function from libMR~\cite{Scheirer_2011_TPAMI}, the MiniBatchKMeans function from scikit-learn~\cite{sculley2010web}, the number of the centre of known categories of subject $N$, quantiles $Q$.
                   \ENSURE The centres of known categories of subject $C$, and libMR models $Model$, the threshold of known categories of subject $Thr$.
                   \STATE $X[i]$ is the abnormal pattern dataset of the subject's $ith$ known categories in which every data $x\in X$ belongs to the subject's $ith$ known categories and is correctly classified by the trained  AI model. $L$ is the number of the subject's known categories.
                   \FOR{$ i=0 \quad to  \quad (L-1)$}
                   \STATE $C[i]=MiniBatchKMeans(X[i],N[i])$
                   \ENDFOR
                   \STATE $Dist=[]$
                   \FOR{$ i=0 \quad to  \quad (L-1)$}
                   \FOR{$ x \quad in  \quad X[i]$}
                   \STATE 
                   $Dist[i]$.add(distance(x,$C[i]$,$C_{others}$) \quad // $distance=sqrt(min\_distance(x,C[i])^2+(1-min\_distance(x,C_{others}))^2)$
                   \ENDFOR
                   \ENDFOR
                   \FOR{$ i=0 \quad to  \quad (L-1)$}
                   \STATE $Model[i]$=FitHigh($Dist[i]$)
                   \STATE $Thr[i]$ is the $Q[i]$ quantile of the $Dist[i]$
                   \ENDFOR
                   \STATE Return $C$, $Model$, $Thr$
         \end{algorithmic}
\end{algorithm}


To adapt our model for open-set recognition of AD diagnosis, the meta-recognition model is used to correct the output of the softmax layer. As shown in Algorithm~\ref{alg3}, OpenAPMax provides a process that modifies the softmax output by using the distance between the abnormal pattern and the centers of different categories. It is an optional step that further enhances the weight of abnormal patterns when identifying samples of unknown categories.


\begin{algorithm}
         \renewcommand{\algorithmicrequire}{\textbf{Input:}}
         \renewcommand{\algorithmicensure}{\textbf{Output:}}
         \caption{\textbf{OpenAPMax probability estimation.}\label{alg3}}
         \begin{algorithmic}[1]
                   \REQUIRE Abnormal pattern of subject $X=\{x_1, x_2, ..., x_n\}$, raw data of subject $Z$, activation vector $V(Z)=\{v_1(Z), v_2(Z)\}$, centres of known categories of subject $C$, libMR models $Model$, threshold of known categories of subject $Thr$, flag $F$, number of "top" classes to revise $\alpha$.
                   \ENSURE The prediction probability $\hat{P}$.
                   \STATE $L$ is the number of the subject's known categories.
                   \STATE Let $s(i)=argsort(v_j(Z))$
                   \STATE Let $Dist=[]$
                   \FOR{$ i=0\quad to  \quad (L-1)$}
                   \STATE $dist[i]=distance(X,C[i],C_{others})$
                   \ENDFOR
                    \FOR{$ i=1\quad to  \quad \alpha$}
                   \STATE $\omega_{i}(Z)=1-\frac{\alpha-i}{\alpha}*Model[i-1].w\_score($dist$[i-1])$
                   \ENDFOR
                   \STATE Revise activation vector $\hat{V}(Z) = V(Z)\circ\omega(Z)$
                   \STATE Define $\hat{v_0}(Z)=\sum_iv_i(Z)(1-\omega_i(Z))$
                   \STATE $\hat{P}(y=j\mid Z)=\frac{e^{\hat{v}_j(Z)}}{\sum_{i=0}^2e^{\hat{v}_i(Z)}}$
                   \IF{$F$}
                    \STATE $abnor\_score=[]$
                    \FOR {$ j=1\quad to  \quad (L-1)$}
                    \STATE $diff=dist[j-1]-Thr[j-1]$
                    \IF{$diff<=0$}
                    \STATE  $abnor\_score$.append(0)
                    \ELSE
                     \STATE $tmp\_abnor\_score=diff/Thr[j-1]$
                     \IF{$tmp\_abnor\_score>1$}
                     \STATE $tmp\_abnor\_score=1$
                     \ENDIF
                      \STATE $abnor\_score$.append($tmp\_abnor\_score$)
                    \ENDIF
                    \ENDFOR
                    \FOR {$ j=1\quad to  \quad (L-1)$}
                     \STATE $\hat{P}(y=j\mid Z)=\hat{P}(y=j\mid Z)*(1-abnor\_score[j-1])$
                    \ENDFOR
                     \STATE$\hat{P}(y=0\mid Z)=1-\sum_{j=1}^{L-1}\hat{P}(y=j\mid Z)$
                    \ENDIF
                   \STATE Return $\hat{P}$
         \end{algorithmic}
\end{algorithm}


\subsection{EVT Modeling}
Extreme value theory is often used to predict the likelihood of small probability events and is an effective method for modeling model prediction scores\cite{10.1117/12.778687, Oza_2019_CVPR}. This approach allows us to estimate the tail probabilities of a random variable beyond a high threshold. The Picklands-Balkema-deHaan formulation, which models probabilities conditioned on random variables exceeding high thresholds, has found widespread application in the field of open set recognition, as exemplified by its use in studies like \cite{Bendale_2016_CVPR} and \cite{yasin2020open}. We focus on determining the distribution function $F_W$ for values of $\omega$ that exceed the threshold $u$. The conditional excess distribution function is then defined as
\begin{equation}
F_U(\omega)=P(\omega - u \leq \omega\mid\omega > u)=\frac{F_W(u + \omega) - F_W(u)}{1 - F_W(u)}
\end{equation}

where, P(·) denotes probability measure function. $F_U$ can be well approximated by the Generalized Pareto Distribution (GPD),

\begin{equation}
G(\omega; \zeta,\mu)=\left\{
\begin{aligned}
  \label{open_set3}
  &1 - (1 + \frac{\zeta \omega}{\mu})^{\frac{1}{\zeta}} \qquad if\  \zeta \neq 0, \\
  &1 - e^{\frac{\omega}{\mu}} \qquad \qquad \quad\ \, if\  \zeta = 0,
\end{aligned}
\right.    
\end{equation}

such that $-\infty < \zeta < +\infty$, $0 < \mu < +\infty$, $\omega > 0$ and $\zeta\omega > -\mu$.

\section{Experiments}
A performance evaluation of OpenAPMax was conducted through three investigations to 1) explore the feasibility of the abnormal pattern on unknown subject identification, 2) demonstrate the superiority of OpenAPMax in different clinical settings, and 3) assess the transplantability of OpenAPMax on different models with different network structures.

\subsection{Dataset}
The data used in the preparation of this article were obtained from the Alzheimer's Disease Neuroimaging Initiative (ADNI) database ( \url{http://adni.loni.usc.edu} ). The ADNI was launched in 2003 as a public-private partnership led by Principal Investigator Michael W. Weiner, MD. For up-to-date information, see \url{http://www.adni-info.org}. The data contain study data, image data, and genetic data compiled by ADNI between 2005 and 2019. Considering the commonly used examinations and the concerned examinations in AD diagnosis by the clinician, the 13 categories of data are selected base information (usually obtained through consultation, including demographics, family history, medical history, etc.), cognition information (usually obtained through consultation and testing, including the Alzheimer's Disease Assessment Scale, Mini-Mental State Exam, Montreal Cognitive Assessment, etc.), cognition testing (usually obtained through testing, including the ANART, Boston Naming Test, Category Fluency-Animals, etc.), neuropsychiatric information (usually obtained through consultation, including the Geriatric Depression Scale, Neuropsychiatric Inventory, etc.), function and behavior information (usually obtained through consultation, including the 
Function Assessment Questionnaire
, Everyday Cognitive Participant Self Report, etc.), physical and neurological examination (usually obtained through testing, including physical characteristics, vitamins, etc.), blood testing, urine testing, nuclear magnetic resonance scan (MRI), positron emission computed tomography scans with 18-FDG, positron emission computed tomography scans with AV45, gene analysis, and cerebral spinal fluid analysis.

In this work, 2127 subjects with 9593 visits are included. To develop the AI model, 85\% AD, and cognitively normal (CN) subjects were divided into the training set, 5\% of the AD and CN subjects were divided into the validation set, and 20\% AD and CN subjects, 100\% MCI subjects, and 100\% SMC subjects were divided into the test set. Note that the MCI and SMC are labeled as unknown to simulate an open setting.

A subject's visit may require different categories of examination. In this work, every combination of those examinations represents a diagnostic strategy. The training set contains 1025 subjects with 3986 visits and generates 180682 strategies according to the collected data. The validation set contains 73 subjects with 254 visits and generates 11898 strategies according to the collected data, and the test set contains 1460 subjects with 5353 visits. In the test set, there were 35 different diagnosis strategies according to different subject situations and 40 different examination abilities of medical institutions.

\subsection{Experimental setup}
Our model was optimized using mini-batch stochastic gradient descent with Adam and a base learning rate of 0.0005~\cite{kingma2015adam}; the model's code will be published to facilitate understanding of the implementation details. All comparison models were constructed and trained according to the needs of AD diagnosis tasks and the settings of their official code. The experiments were conducted on a Linux server equipped with Tesla P40 and Tesla P100 GPUs.
\subsection{Abnormal pattern}
According to the AD diagnosis guidelines and recommendations from the expert group in ADNI, as shown in Table~\ref{index_nomal}, 14 indicators are selected. Among the indicators shown in Table~\ref{index_nomal}, except for the MMSE and MOCA, which are able to directly find the normal range of the AD category and CN category in the AD diagnostic guide, the normal range of other indicators is determined by statistics. In this work, for a category, we sort the indicator values, and the normal range for a category is obtained by the 5th and 95th percentiles of the indicator values' distribution.

\begin{table*}
\centering
\caption{\textbf{The normal range of indicators.}\label{index_nomal}}

\begin{tabular}{cccccc}
\hline
\multicolumn{2}{c}{\multirow{2}{*}{}}                                                                                                                                         & \multicolumn{2}{c}{AD Normal} & \multicolumn{2}{c}{CN Normal} \\
\multicolumn{2}{c}{}                                                                                                                                                          & Low            & High         & Low           & High          \\
\hline
\multirow{2}{*}{\begin{tabular}[c]{@{}c@{}}\\Medical\\ history\end{tabular}}                             & Psychiatric                                                          & 0              & 0            & 0             & 0             \\
                                                                                                       & \begin{tabular}[c]{@{}c@{}}Neurologic\\ (other than AD)\end{tabular} & 0              & 0            & 0             & 0             \\
\multirow{2}{*}{Symptoms\footnotemark[1]}                                                                              & Present\_count\_21\footnotemark[2]                                                   & 0              & 6            & 0             & 6             \\                                    & Present\_count\_28\footnotemark[3]                                                   & 0              & 8            & 0             & 8             \\
\multirow{2}{*}{\begin{tabular}[c]{@{}c@{}}Cognitive\\ Change Index\footnotemark[4]\end{tabular}}                      & Score\_12\footnotemark[5]                                                            & 32.2188        & 60           & 12            & 13.5634       \\
                                                                                                       & Score\_20\footnotemark[6]                                                            & 50.3438        & 100          & 20            & 22.0845       \\
CDRSB\footnotemark[7]                                                                                                  &                                                                      & 2              & 18           & 0             & 0             \\
\multirow{3}{*}{\begin{tabular}[c]{@{}c@{}}Alzheimer's Disease \\ Assessment Scale\footnotemark[8]\end{tabular}}       & ADAS11\footnotemark[9]                                                               & 10             & 70           & 0             & 11.264        \\
                                                                                                       & ADAS13\footnotemark[10]                                                               & 18             & 85           & 0             & 17.67         \\
                                                                                                       & ADASQ4                                                               & 5              & 10           & 0             & 6             \\
MMSE\footnotemark[11]                                                                                                   &                                                                      & 0              & 27           & 25            & 30            \\
MOCA\footnotemark[12]                                                                                                   &                                                                      & 0              & 23           & 26            & 30            \\
\multirow{2}{*}{\begin{tabular}[c]{@{}c@{}}Preclinical Alzheimer's\\Cognitive Composite\footnotemark[13]\end{tabular}} & mPACCdigit                                                           & -30.0745       & -7.6955      & -5.1733       & 4.7304        \\
                                                                                                       & mPCCtrailsB                                                          & -29.7277       & -6.7798      & -4.8523       & 4.3338 \\
                                                                                
\hline                                                                                     
\end{tabular}

\end{table*}
\footnoterule
\footnotetext[1]{Nausea, Vomiting, Diarrhoea, Constipation, Abdominal discomfort, Sweating, Dizziness, Low energy, Drowsiness, Blurred vision, Headache, Dry mouth, Shortness of breath, Coughing, Palpitations, Chest pain, Urinary discomfort (e.g., burning), Urinary frequency, Ankle swelling, Musculoskeletal pain, Rash, Insomnia, Depressed mood, Crying, Elevated mood, Wandering, Fall, Other.}
\footnotetext[2]{Nausea to Rash}
\footnotetext[3]{Nausea to Other}
\footnotetext[4]{The CCI scale can be found at \url{https://adni.bitbucket.io/reference/cci.html}.}
\footnotetext[5]{CCI1 to CCI12}
\footnotetext[6]{CCI1 to CCI20}
\footnotetext[7]{The CDR scale can be found at \url{https://adni.bitbucket.io/reference/cdr.html}.}
\footnotetext[8]{The Alzheimer's Disease Assessment Scale-Cognitive scale can be found at\url{https://adni.bitbucket.io/reference/adas.html}.}
\footnotetext[9]{Q1 to Q11}
\footnotetext[10]{Q1 to Q13}
\footnotetext[11]{The Mini Mental State Exam scale can be found at \url{https://adni.bitbu\\cket.io/reference/mmse.html}.}
\footnotetext[12]{The Montreal Cognitive Assessment scale can be found at\url{https://adni.bitbu\\cket.io/reference/moca.html}.}
\footnotetext[13]{The calculation method of the Preclinical Alzheimer's Cognitive Composite can be found at \url{https://ida.loni.usc.edu/pages/access/studyData.jsp?\\categoryId=16&subCategoryId=43}.}

\begin{figure*}
\begin{minipage}[htbp]{1\linewidth}
\subfloat[]{
\hspace{1cm}\includegraphics[width=0.4\linewidth]{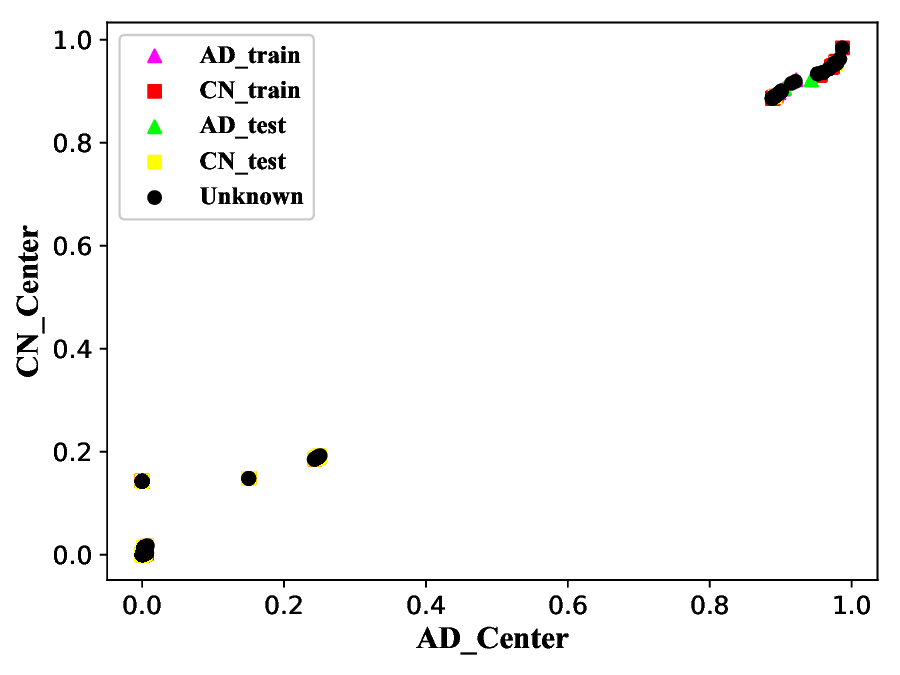}
}
\quad
\subfloat[]{
\hspace{-0.3cm}\includegraphics[width=0.4\linewidth]{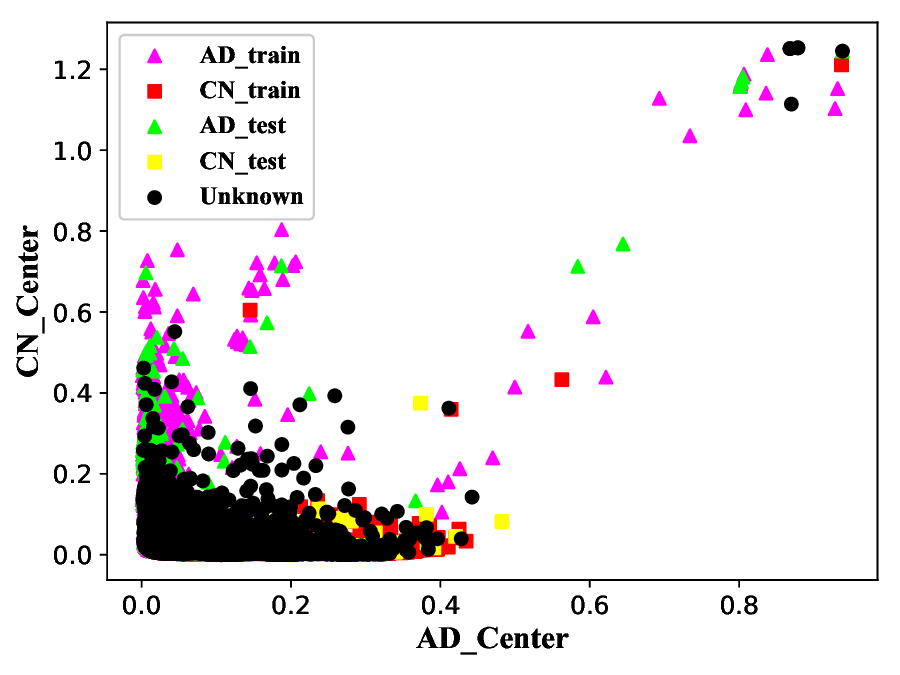}
}
\end{minipage}

\begin{minipage}[h]{1\linewidth}

\subfloat[]{
\centerline{
\hspace{-0.8cm}\includegraphics[width=0.7\linewidth]{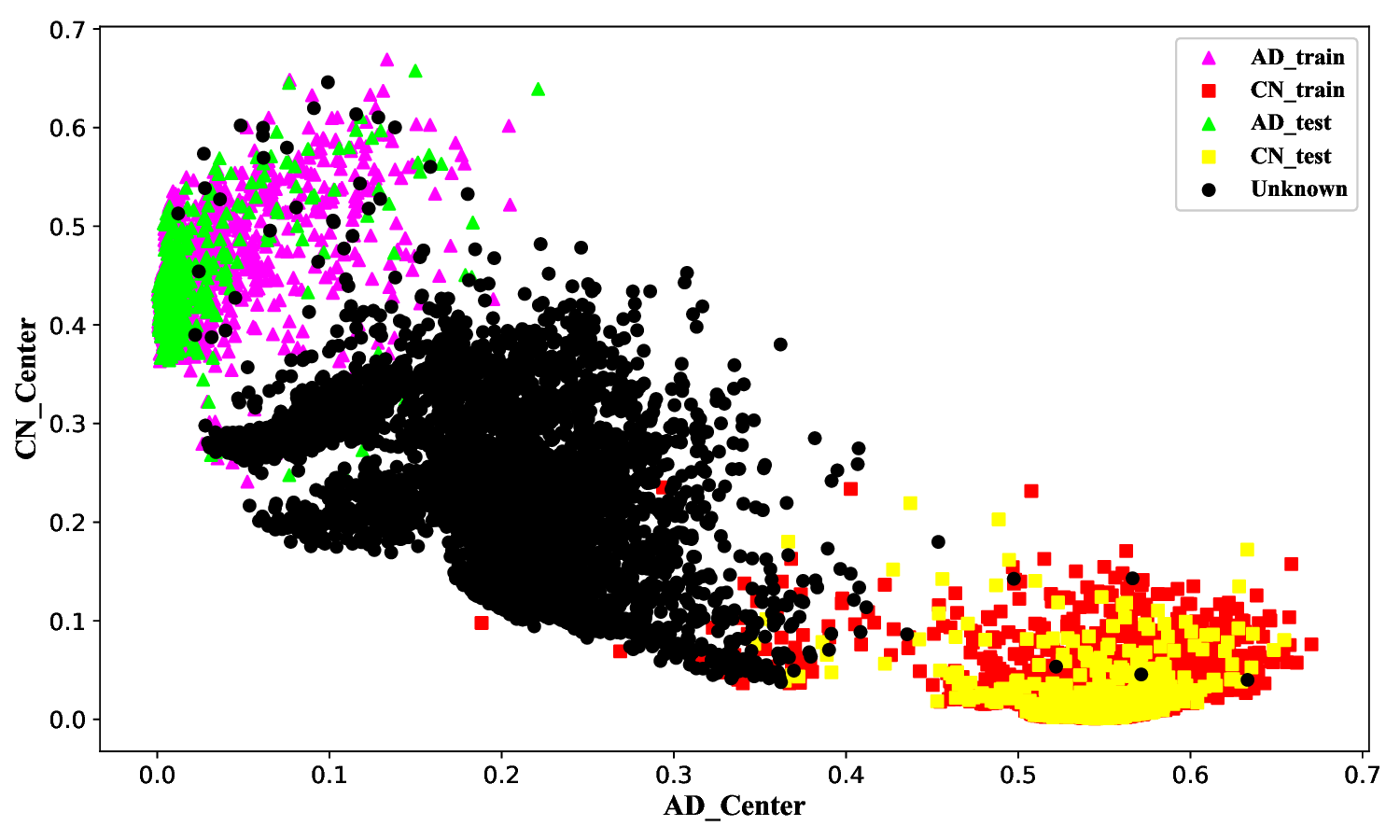}
}}
\end{minipage}

\caption{Comparison of unknown sample recognition ability. For all figures, the abscissa is the distance between the sample and the AD category center, and the ordinate is the distance between the sample and the CN category. (a) Each sample is composed of basic information and routine cognitive test data, which are easily obtained at the time of the visit for every subject. (b) Each sample is composed of 14 selected indicators from basic information and routine cognitive test data. (c) Each sample is composed of an abnormal pattern, which is derived from the 14 selected indicators. \label{ap}}
\label{example}
\end{figure*}

To investigate the role of abnormal patterns in identifying unknown samples in AD diagnosis, we compared the identification of unknown samples by using the similarity of patient data and the identification of unknown samples by using the familiarity of patients' abnormal patterns.

First, we directly used all of the patients' basic information and cognitive testing information to calculate the similarity between samples and the similarity between samples and the known category centers. As shown in Figure~\ref{ap}(a), we find that the distribution of unknown samples highly overlaps with that of known samples (AD and CN), and it is difficult to separate unknown samples from known samples. The reason for this phenomenon is that unknown samples contain a large number of MCI patients, while MCI patients are highly similar to AD patients. The boundary between MCI patients and AD patients is blurred, and some MCI patients will become AD patients in the near future.   

Second, we used the 14 indicator values selected from the basic information and cognitive testing information above to calculate the similarity between samples and the similarity between samples and known category centers. As shown in Figure~\ref{ap}(b), although the distribution of unknown samples is obviously different from that of known samples, most unknown samples still overlap with known samples. The process of selecting 14 indicators can be considered feature selection in machine learning, which can help improve the performance of model classification. However, it is not very helpful for identifying unknown samples.

Finally, according to the 14 selected indicators above and the normal indicator range of each category, we obtained the abnormal pattern of each sample and used it to calculate the similarity between samples and the similarity between samples and known category centers. As shown in Figure~\ref{ap}(c), the samples of known categories are distributed in the upper left and lower right corners of the figure. At the same time, the remaining space is occupied by samples of unknown categories. Although samples of known categories and samples of unknown categories still overlap, most samples of unknown categories have obvious differences from known samples, which means that abnormal patterns are important information for identifying unknown patients.

\subsection{Methods for comparison}
We validate the effectiveness of OpenAPMax by comparing it with related works: the discriminative models, OpenMax (2016)~\cite{bendale2016towards} and OVRN (2022)~\cite{jang2022collective}, and the generative models, Gen-Dis (2020)~\cite{perera2020generative}, in different clinical settings. In addition, we also compared the performance of OpenAPMax with the related work image-based models (DSA-3D-CNN (2016)~\cite{hosseini2016alzheimer}, CNN-LRP (2019)~\cite{bohle2019layer}, VoxCNN-ResNet (2017)~\cite{korolev2017residual}, Dynamic-VGG (2020)~\cite{xing2020dynamic}) and the multimodal inputs model (FCN-MLP (2020)~\cite{qiu2020development}).

\begin{table*}[htbp]
  \centering
  \caption{Comparison of OpenAPMax with different models in different clinical settings.}\label{setting_performance}
    \resizebox{\linewidth}{!}{
    \begin{tabular}{cllllll}
    \toprule
    \multicolumn{2}{c}{\multirow{2}[2]{*}{Model}} & \multicolumn{2}{c}{AUC(95\% CI\footnotemark[1])} & \multicolumn{3}{c}{Sensitivity(95\% CI)} \\
    \multicolumn{2}{c}{} & \multicolumn{1}{c}{AD} & \multicolumn{1}{c}{CN} & \multicolumn{1}{c}{AD} & \multicolumn{1}{c}{CN} & \multicolumn{1}{c}{Unknown} \\
    \midrule\midrule
    \multirow{3}[1]{*}{OpenMax~\cite{bendale2016towards}} & Sc\_Dataset & 0.7541(0.7091-0.7935) & 0.6252(0.5971-0.6531) & 0.7591(0.6826-0.8267) & 0.1968(0.1442-0.2539) & 0.4471(0.4265-0.4680) \\
          & S\_Dataset & 0.7728(0.7299-0.8112) & 0.7238(0.6954-0.7521) & 0.5468(0.4672-0.6278) & 0.7789(0.7184-0.8367) & 0.3433(0.3236-0.3633) \\
          & M\_Dataset & 0.8558(0.8331-0.8772) & 0.8270(0.8036-0.8519) & \textbf{0.9321(0.8889-0.9686)} & 0.8483(0.7959-0.8980) & 0.2863(0.2672-0.3045) \\
    \midrule
    \multirow{3}[1]{*}{Gen-Dis~\cite{perera2020generative}} & Sc\_Dataset & 0.8082(0.7771-0.8396) & 0.7534(0.7267-0.7795) & 0.7484(0.6752-0.8170) & 0.7173(0.6538-0.7778) & 0.2574(0.2382-0.2771) \\
          & S\_Dataset & 0.8267(0.7965-0.8562) & 0.7115(0.6815-0.7397) & 0.7483(0.6715-0.8203) & 0.7111(0.6436-0.7720) & 0.2391(0.2211-0.2571) \\
          & M\_Dataset & 0.8565(0.8336-0.8768) & 0.8280(0.8012-0.8512) & 0.7081(0.6363-0.7778) & 0.5872(0.5187-0.6533) & 0.3282(0.3096-0.3471) \\
    \midrule
    \multirow{3}[1]{*}{OVRN~\cite{jang2022collective}} & Sc\_Dataset & 0.7570(0.7186-0.7921) & 0.7045(0.6713-0.7414) & 0.7517(0.6757-0.8201) & 0.6703(0.6032-0.7347) & 0.4709(0.4491-0.4938) \\
          & S\_Dataset & 0.7684(0.7288-0.8044) & 0.6974(0.6620-0.7342) & 0.7500(0.6735-0.8205) & 0.5951(0.5254-0.6667) & 0.5557(0.5343-0.5774) \\
          & M\_Dataset & 0.8340(0.8011-0.8637) & 0.6636(0.6291-0.6976) & 0.8284(0.7654-0.8844) & 0.6231(0.5579-0.6884) & 0.5095(0.4884-0.5291) \\
    \midrule
    \multirow{3}[0]{*}{\textbf{OpenAPMax}} & Sc\_Dataset & 0.9508(0.9248-0.9717) & 0.9913(0.9815-0.9979) & 0.5655(0.4797-0.6463) & 0.8556(0.8028-0.9010) & \textbf{0.9932(0.9897-0.9963)} \\
          & S\_Dataset & \textbf{0.9653(0.9524-0.9752)} & \textbf{0.9941(0.9868-0.9990)} & 0.7385(0.6626-0.8029) & \textbf{0.8865(0.8381-0.9277)} & 0.9394(0.9290-0.9496) \\
          & M\_Dataset & 0.9502(0.9304-0.9662) & 0.9927(0.9854-0.9981) & 0.8492(0.7891-0.9051) & 0.8127(0.7551-0.8667) & 0.9396(0.9290-0.9492) \\
    \toprule
    \end{tabular}%
    }
  \label{tab:addlabel}%
  \begin{tablenotes}
     \item[1] CI = confidence interval. To evaluate the AI model's evaluation index, a non-parametric bootstrap method is applied to calculate the CI for the evaluation index~\cite{efron1994introduction}. In this work, we calculate 95\% CI for every evaluation index, and we randomly sampled $2500$ cases from the test set and evaluated the AI model by the sampled set for every evaluation index. $2000$ repeated trials are executed, and $2000$ evaluation index values are generated. The 95\% CI is obtained by the 2.5 and 97.5 percentiles of the evaluation index values' distribution.
   \end{tablenotes}
\end{table*}%

\begin{table*}[htbp]
  \centering
  \caption{OpenMax vs. OVRN vs. OpenAPMax with different models.}\label{real_model_performance}
  \resizebox{\linewidth}{!}{
    \begin{tabular}{cllllll}
    \toprule
    \multicolumn{2}{c}{\multirow{2}[1]{*}{Model}} & \multicolumn{2}{c}{AUC(95\% CI)} & \multicolumn{3}{c}{Sensitivity(95\% CI)} \\
    \multicolumn{2}{c}{} & \multicolumn{1}{c}{AD} & \multicolumn{1}{c}{CN} & \multicolumn{1}{c}{AD} & \multicolumn{1}{c}{CN} & \multicolumn{1}{c}{Unknown} \\
    \midrule\midrule
    \multirow{3}[0]{*}{CNN-LRP~\cite{bohle2019layer}} & OpenMax & 0.8300(0.7921-0.8641) & 0.6794(0.6471-0.7080) & 0.8486(0.7852-0.9044) & 0.4875(0.4144-0.5549) & 0.2928(0.2740-0.3130) \\
          & OVRN  & 0.7082(0.6650-0.7502) & 0.6148(0.5763-0.6529) & 0.5274(0.4422-0.6096) & 0.5393(0.4681-0.6111) & 0.5543(0.5332-0.5756) \\
          & \textbf{OpenAPMax} & \textbf{0.9277(0.8887-0.9602)} & \textbf{0.8967(0.8593-0.9293)} & \textbf{0.8605(0.7985-0.9160)} & \textbf{0.7241(0.6558-0.7895)} & \textbf{0.9926(0.9886-0.9958)} \\
    \midrule
    \multirow{3}[1]{*}{DSA-3D-CNN~\cite{hosseini2016alzheimer}} & OpenMax & 0.7779(0.7382-0.8147) & 0.6685(0.6349-0.7038) & 0.3643(0.2824-0.4468) & 0.3415(0.2743-0.4118) & 0.7105(0.6909-0.7302) \\
          & OVRN  &   0.6467(0.6074-0.6893) & 0.5280(0.5032-0.5580) &   0.3718(0.2946-0.4527) &  0.1622(0.1129-0.2176) & 0.8038(0.7871-0.8201) \\
          & \textbf{OpenAPMax} & \textbf{0.9011(0.8640-0.9338)} & \textbf{0.9588(0.9346-0.9771)} & \textbf{0.6870(0.6061-0.7613)} & \textbf{0.8528(0.8033-0.9018)} & \textbf{0.9749(0.9680-0.9811)} \\
    \midrule
    \multirow{3}[1]{*}{
VoxCNN-ResNet
~\cite{korolev2017residual}} & OpenMax & 0.7616(0.7212-0.8006) & 0.6375(0.6010-0.6725) & 0.2349(0.1625-0.3077) & 0.3182(0.2500-0.3850) & 0.6973(0.6768-0.7169) \\
          & OVRN  & 0.6744(0.6308-0.7143) & 0.5668(0.5315-0.6050) & 0.4765(0.3910-0.5538) & 0.4354(0.3656-0.5084) & 0.5447(0.5240-0.5657) \\
          & \textbf{OpenAPMax} & \textbf{0.9365(0.9064-0.9617)} & \textbf{0.9801(0.9667-0.9907)} & \textbf{0.7500(0.6737-0.8188)} & \textbf{0.8172(0.7586-0.8721)} & \textbf{0.9949(0.9913-0.9977)} \\
    \midrule
    \multirow{3}[0]{*}{FCN-MLP~\cite{qiu2020development}} & OpenMax & 0.9255(0.9032-0.9448) & 0.7424(0.7137-0.7711) & 0.7209(0.6444-0.7941) & 0.6633(0.5978-0.7330) & 0.5822(0.5624-0.6021) \\
          & OVRN  & 0.8306(0.7932-0.8643) & 0.6972(0.6674-0.7261) & 0.7746(0.7007-0.8418) & 0.8391(0.7824-0.8906) & 0.4044(0.3835-0.4255) \\
          & \textbf{OpenAPMax} & \textbf{0.9577(0.9310-0.9796)} & \textbf{0.9811(0.9656-0.9930)} & \textbf{0.8750(0.8176-0.9279)} &  \textbf{0.9296(0.8894-0.9636)} & \textbf{0.9963(0.9935-0.9986)} \\
    \midrule
    \multirow{3}[1]{*}{Dynamic-VGG~\cite{xing2020dynamic}} & OpenMax & 0.7150(0.6719-0.7557) & 0.6329(0.5959-0.6668) & 0.3788(0.3047-0.4604) & 0.1568(0.1067-0.2111) & 0.6887(0.6685-0.7080) \\
          & OVRN  & 0.6189(0.5784-0.6592) & 0.5676(0.5339-0.6057) & 0.3631(0.2817-0.4444) & 0.3121(0.2455-0.3806) & 0.6829(0.6622-0.7020) \\
          & \textbf{OpenAPMax} & \textbf{0.8998(0.8544-0.9395)} & \textbf{0.8878(0.8551-0.9168)} & \textbf{0.8023(0.7338-0.8647)} & \textbf{0.6237(0.5512-0.6923)} & \textbf{0.9935(0.9898-0.9963)} \\
    \bottomrule
    \end{tabular}%
    }
\end{table*}%

\subsection{Performances in different methods}
To investigate the performance of OpenAPMax, we built a simple screening dataset $Sc\_{Dataset}$, a single diagnostic strategy dataset $S\_Dataset$ and a multiple diagnostic strategy dataset $M\_Dataset$. For $Sc\_{Dataset}$, every subject only contains basic information and cognitive testing information. $Sc\_{Dataset}$ can be collected outside of the hospital. For $S\_Dataset$, every subject contains basic information, cognitive testing information, and MRI. $S\_Dataset$ construction rules are the dataset construction rules that most of the recent research has followed. $M\_Dataset$ contains 13 categories of commonly used information during AD diagnosis. and every sample contains one or more categories according to the diagnosis strategies. Four models, the OpenMax-based discriminative model, the generative model, the OVRN model, and the OpenAPMax-based discriminative model, were run on $Sc\_{Dataset}$, $S\_Dataset$ and $M\_Dataset$ for comparison.

As shown in Table~\ref{setting_performance}, OpenAPMax achieved the best performance in out-of-hospital screening, the traditional single-diagnosis strategy setting, and the real-world multi-diagnosis strategy clinical setting. The performance of OpenMax and the generative model Gen-Dis are poor in different settings, especially in real-world multi-strategy settings. Note that OpenMax, Gen-Dis, and OVRN all sacrifice the ability to recognize unknown samples (0.29, 0.33, and 0.51) in exchange for the ability to recognize known samples in a real-world setting. The reasons for this phenomenon may be as follows: (1) The subjects' basic information and cognitive information are subjective, and these data may present a complex distribution according to different patients and different collectors; (2) MRI data from more than 60 institutions with different instruments makes the data distribution very complex; and (3) the training set has more than 4000 different diagnostic strategies, which further increases the complexity of the data distribution. Comparing the performance of different models in different clinical settings, the use of multimodal data and multi-strategy combination combinations can enhance the models performance. However, compared with AD diagnosis in the closed world, AD diagnosis in the real world still has huge room for improvement, and further exploration is needed to promote the implementation of AD diagnosis in real clinical practice ~\cite{tanveer2020machine,mahajan2020machine}.

\subsection{OpenAPMax vs. OVRN vs. OpenMax with different models}

To investigate the transferability of OpenAPMax, we applied OpenMax, OVRN and OpenAPMax to several AD diagnostic models that were trained on closed sets.

As shown in Table~\ref{real_model_performance}, for all compared models, OpenAPMax demonstrates great advantages over OpenMax and OVRN. Compared with AD subjects, all models with OpenAPMax have better recognition performance for CN subjects. Furthermore, except for Dynamic-VGG-OpenAPMax, the AUC scores of all other compared models are larger than 0.9. The AUC scores of CN subjects in the results generated by the OpenAPMax method are generally higher than those of AD. The reason for the above phenomenon may be that subjects of unknown categories are more similar to AD subjects but different from CN subjects. Thus, CN subjects are easy to identify. Compared with OpenMax and OVRN, OpenAPMax exhibits excellent sensitivity to subjects of unknown categories and maintains acceptable sensitivity for subjects of AD and CN, which proves once again that abnormal patterns play an important role in excluding unknown subjects.

\subsection{Potential clinical applications}
In real-world clinical settings, both the subject and medical institution are complex and various. There is no one-size-fits-all strategy for diagnosing every subject in every medical institution. For example, AD subjects usually need a nuclear magnetic resonance scan, while CN subjects usually do not, and subjects in hospitals in developed regions may receive a PET examination, while those in non-developed regions do not suffer due to the lack of PET equipment in hospitals.

In this work, to model the AD diagnosis task in a real-world setting, the classification model simultaneously modelled the samples under different diagnosis strategies. In the training set, 180682 diagnosis strategies generated 180682 samples for model training. Note that there are 4096 different diagnosis strategies in the training set since similar subjects may share the same diagnosis strategy. In the test set, 1460 subjects with 5353 visits were diagnosed 
by 5353 strategies that correspond to 35 different diagnosis strategies
. The training set contains almost all possible AD diagnosis strategies, while the test set contains all commonly used AD diagnostic strategies. Compared with the fixed diagnosis strategy model, our model has great potential to be applied in complex and changeable clinical settings.

Although the accuracy of AD diagnosis in the closed-set setting is close to 100\%, the accuracy of AD diagnosis in the open-set environment is only approximately 80\%~\cite{tanveer2020machine,mahajan2020machine}. At present, it seems that the AD diagnostic model based on open-set recognition has a high probability of misdiagnosis, and we need to further improve the model's performance. However, there is no clear consensus on the model's accuracy before it can be applied to the real-world clinical setting. Thus, it is also an urgent problem to explore the standard of the AD model in clinical settings.

\section{Conclusion}
The first contribution of this paper is that it provides a new perspective for identifying unknown subjects from known subjects; this perspective integrates the prior knowledge of clinicians through abnormal patterns. We propose the abnormal-pattern-based open-set recognition mechanism OpenAPMax, which considers the abnormal pattern of subjects rather than the similarity shown by subjects' raw data. The experimental results demonstrate that OpenAPMax directly yields state-of-the-art performance on AD diagnosis in different clinical settings. Furthermore, the experiment shows high portability in that OpenAPMax is able to integrate into other models without changing their architecture. Finally, we expect that subjects' abnormal patterns will serve as useful information for developing other disease diagnosis models in an open-set setting or for related analysis in future works, similar to the role it plays in this paper.


%



\ifCLASSOPTIONcompsoc
  \section*{Acknowledgements}
\else
  \section*{Acknowledgement}
\fi

This work is supported by the Project of the National Natural Science Foundation of China (Grant No. 61967002), the Project of Guangxi Science and Technology (Grant No. GuiKeAD20297004), and the Key Program of the National Natural Science Foundation of China (Grant No. U21A20474).

\ifCLASSOPTIONcaptionsoff
  \newpage
\fi




\bibliographystyle{IEEEtran}
\bibliography{main}




%

\begin{IEEEbiography}[{\includegraphics[width=1in,height=1.25in,clip,keepaspectratio]{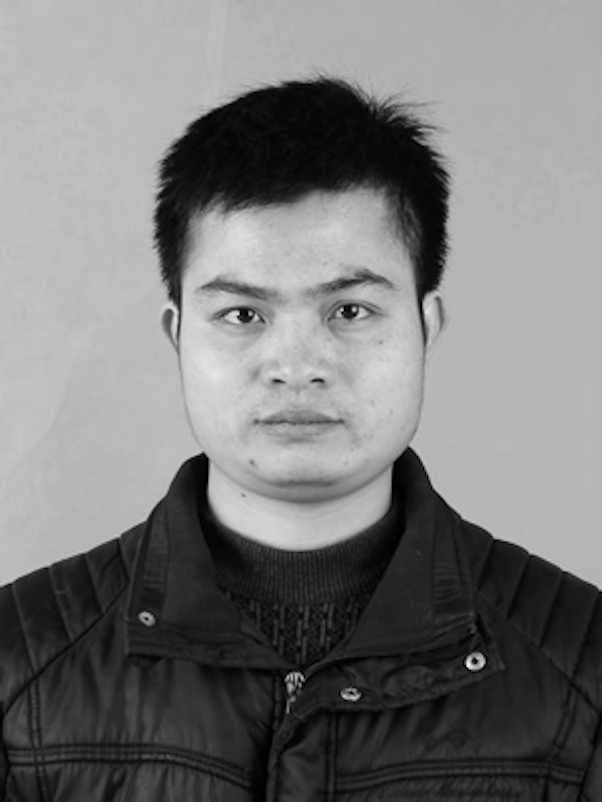}}]{Yunyou Huang}
received a B.S. degree and an M.S. degree
from Guangxi Normal University in 2012 and 2015, respectively,
and a PhD degree in computer software and theory from the
University of Chinese Academy of Sciences in 2020. He has been an assistant professor in computer
science at Guangxi Normal University since 2020. His
research interests focus on big data and machine
learning.
\end{IEEEbiography}

\begin{IEEEbiography}[{\includegraphics[width=1in,height=1.25in,clip,keepaspectratio]{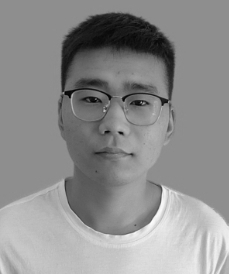}}]{Xianglong Guan}
received a B.S. degree from the Tongda College of Nanjing University of Posts and Telecommunications in 2020. He is now studying for an M.S. degree at the School of Electronics and Information Engineering \& Integration School of Guangxi Normal University. His research interests focus on artificial medical intelligence and open-set recognition.
\end{IEEEbiography}

\begin{IEEEbiography}[{\includegraphics[width=1in,height=1.25in,clip,keepaspectratio]{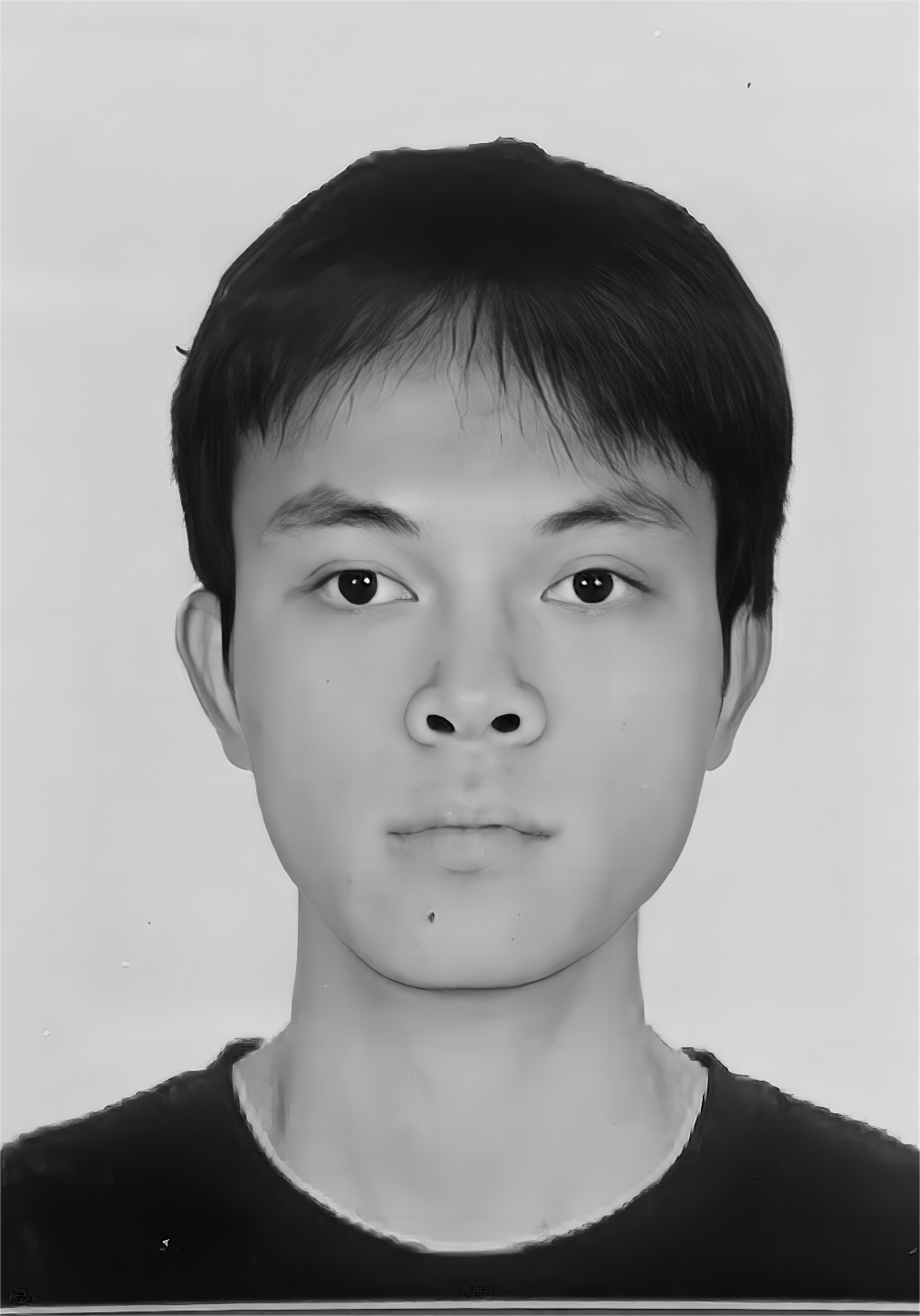}}]{Xiangjiang Lu}received a B.S. degree in Software Engineering from the Guilin University of Technology, China, in 2021. He is currently pursuing an M.S. degree at the School of Computer Science and Engineering \& School of Software, Guangxi Normal University, China. His research interests include clinical AI and medical image analysis.
\end{IEEEbiography}

\begin{IEEEbiography}[{\includegraphics[width=1in,height=1.25in,clip,keepaspectratio]{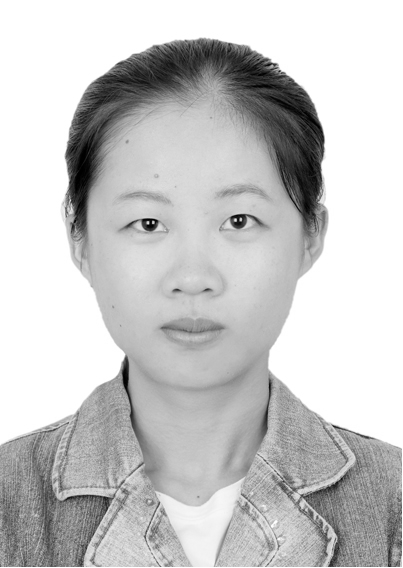}}]{Xiaoshuang Liang}received a B.S. degree in Computer Science and Technology from Guangxi Normal University, China, in 2021. She is currently pursuing the Master degree at the School of Computer Science and Engineering \& School of Software, Guangxi Normal University, China. Her research interests include Clinical AI, Interpretability of Clinical Medicine.
\end{IEEEbiography}

\begin{IEEEbiography}[{\includegraphics[width=1in,height=1.25in,clip,keepaspectratio]{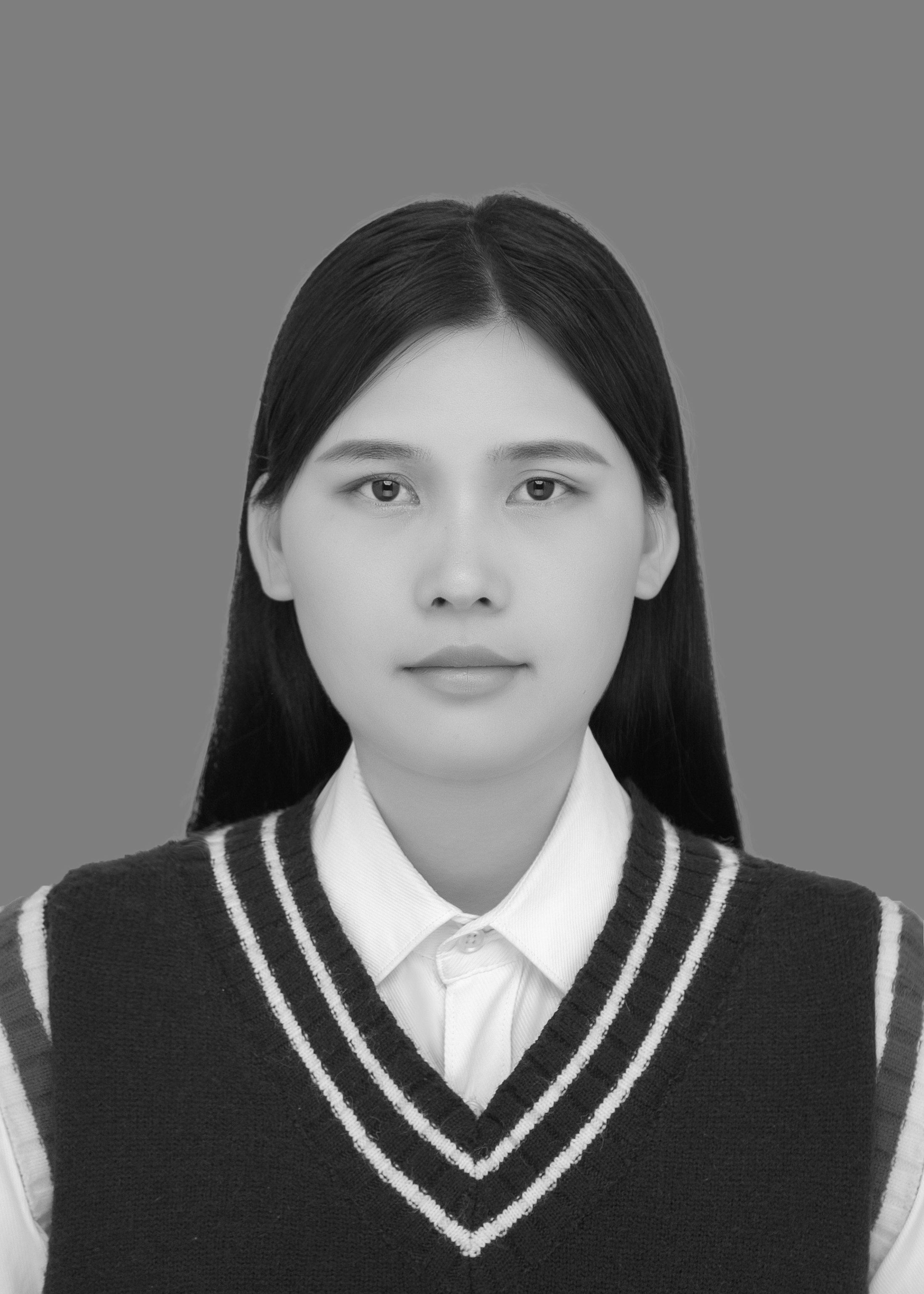}}]{Xiuxia Miao}received a B.S. degree in Software Engineering from Hezhou University, China, in 2020. She is currently pursuing a Master's degree at the School of Computer Science and Engineering \& School of Software, Guangxi Normal University, China. Her research interests include 
clinical benchmark and clinical benchmark test systems.

\end{IEEEbiography}

\begin{IEEEbiography}[{\includegraphics[width=1in,height=1.25in,clip,keepaspectratio]{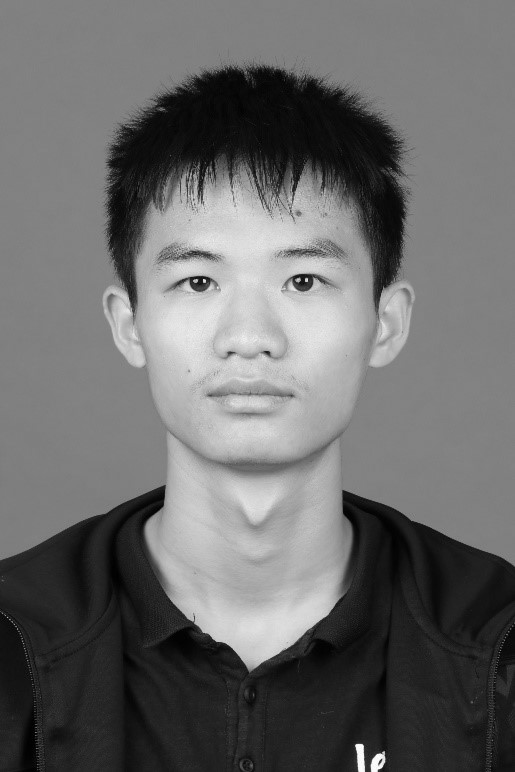}}]{Jiyue Xie}received a B.S. degree from the Guangxi Normal University for Nationalities in July 2022 and is currently studying for a Master of Software Engineering at Guangxi Normal University. His research interests include deep learning.
\end{IEEEbiography}

\begin{IEEEbiography}[{\includegraphics[width=1in,height=1.25in,clip,keepaspectratio]{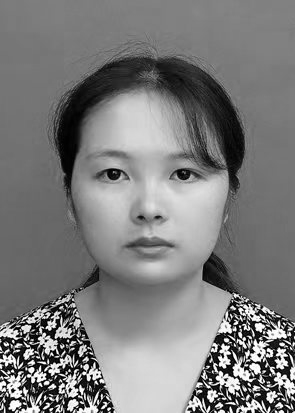}}]{Wenjing Liu}received a B.S. degree in Software Engineering from the Hunan Institute of Science and Technology, China, in 2017. She is pursuing her M.S. degree at the Computer Science and Engineering \& School of Software at Guangxi Normal University, China. Her research interests include clinical AI and virtual patients.
\end{IEEEbiography}

\begin{IEEEbiography}[{\includegraphics[width=1in,height=1.25in,clip,keepaspectratio]{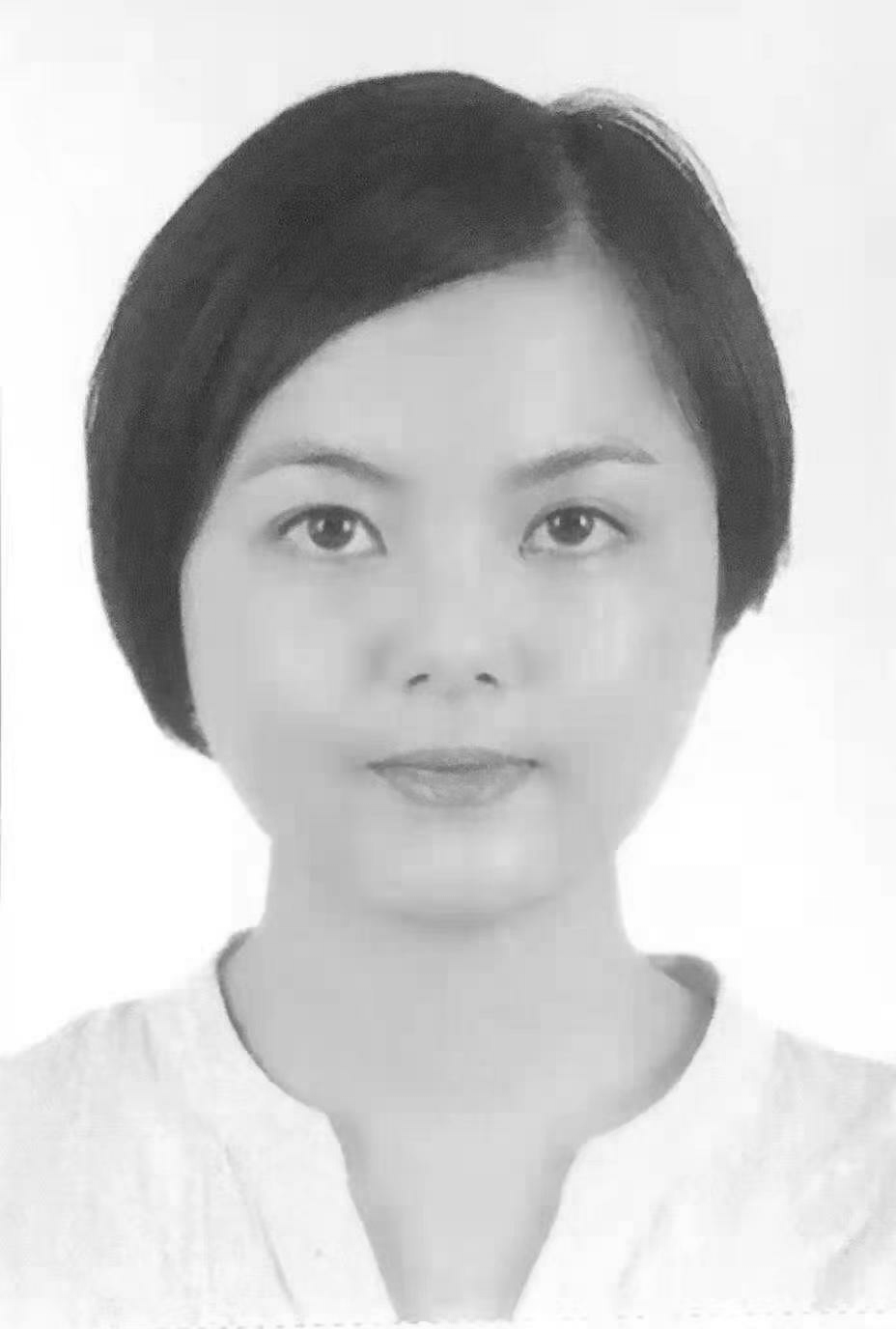}}]{Li Ma}received an M.S. degree in computer software and theory in 2009. She is currently a full-time computer teacher at Guilin Medical University. Her current research interests include medical information processing, knowledge graphs, and artificial intelligence ethics.
\end{IEEEbiography}

\begin{IEEEbiography}[{\includegraphics[width=1in,height=1.25in,clip,keepaspectratio]{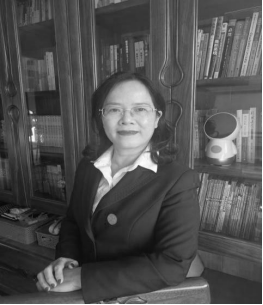}}]{Suqin Tang}, PhD and Professor, received a B.S. and M.S. degrees from Guangxi Normal University,
Guilin, China. In addition, she received a PhD degree from Central South University, Changsha, China. Her main research interests include ontology, description logic, knowledge engineering, and intelligent tutoring systems.
\end{IEEEbiography}

\begin{IEEEbiography}[{\includegraphics[width=1in,height=1.25in,clip,keepaspectratio]{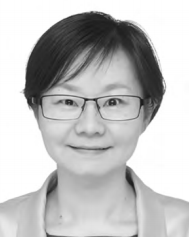}}]{Zhifei Zhang}
received an M.S. degree in physiology and a PhD degree in aetiology from Capital Medical University, Beijing, China, in 2003 and
2011, respectively, where she is currently an Associate Professor with the Beijing Key Laboratory of
Respirology. Her research interests include medical big data and circulatory physiology.
\end{IEEEbiography}

\begin{IEEEbiography}[{\includegraphics[width=1in,height=1.25in,clip,keepaspectratio]{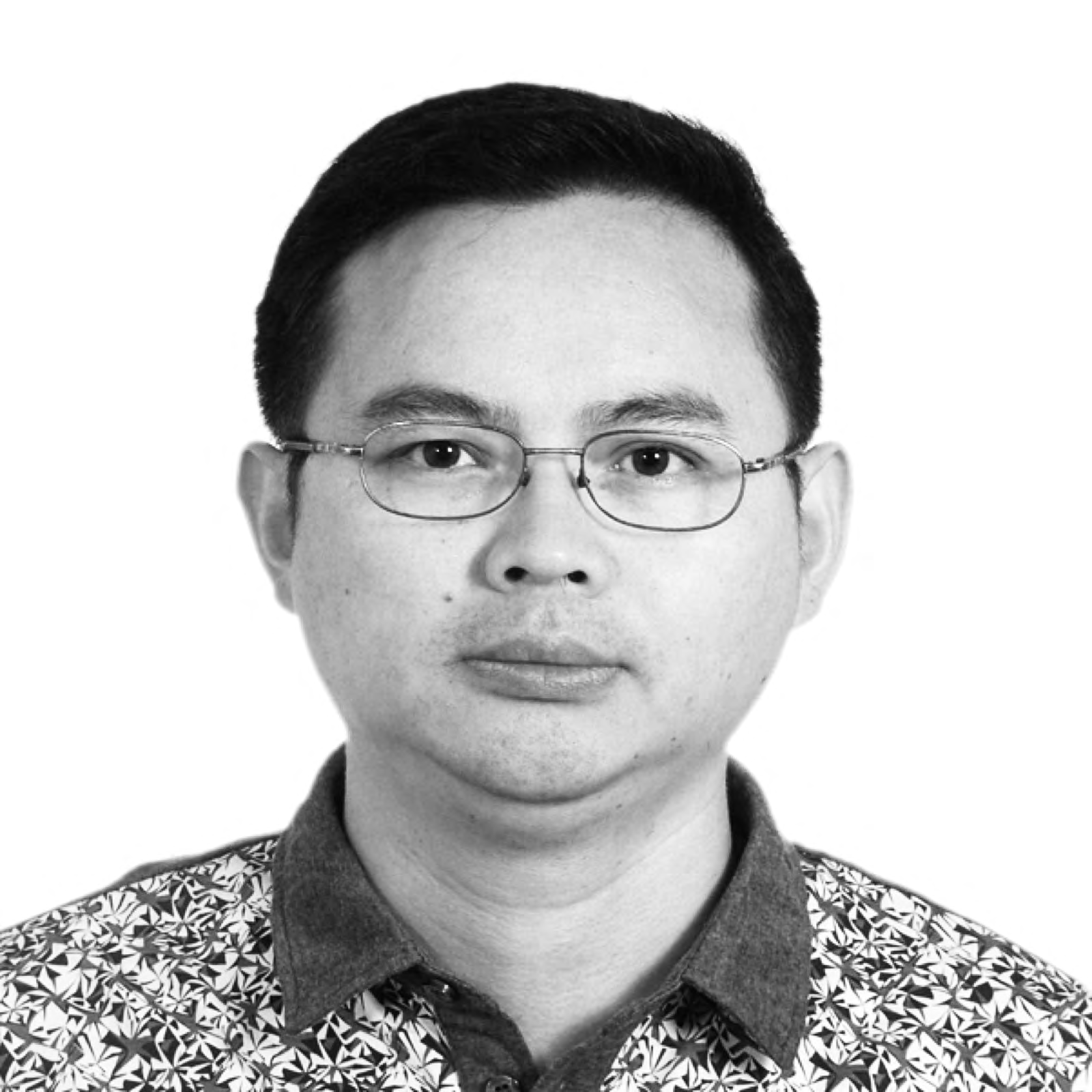}}]{Jianfeng Zhan}is a Full Professor at the Institute of
Computing Technology (ICT), Chinese Academy of Sciences (CAS), and University of Chinese Academy of Sciences (UCAS). He is also the director of the Software Systems
Labs, ICT, CAS. He has supervised over 90 graduate
students, postdocs, and engineers in the past two
decades. His research areas span from chips and systems to
benchmarks. A common thread is benchmarking,
designing, implementing, and optimizing parallel
and distributing systems. He has made strong and
effective efforts to transfer his academic research into
advanced technology to impact general-purpose production systems. Several technical innovations and research results, including 36
patents from his team have been widely adopted in benchmarks, operating systems
and cluster and cloud system software with direct contributions to the advancement of parallel and distributed systems in China and throughout the world. Dr.
Jianfeng Zhan founded and chairs BenchCouncil, and he has served as the IEEE TPDS Associate
Editor since 2018. He received the second-class Chinese National Technology Promotion Prize in 2006, the Distinguished Achievement Award of the Chinese Academy of Sciences in 2005, and the IISWC best paper award in 2013. Jianfeng
Zhan received his B.E. in Civil Engineering and MSc in Solid Mechanics from
Southwest Jiaotong University in 1996 and 1999, and his PhD in Computer Science
from the Institute of Software, CAS and UCAS in 2002.
\end{IEEEbiography}





\end{document}